# Topography scanning as a part of process monitoring in power cable insulation process

Janne HARJUHAHTO, Jaakko HARJUHAHTO, Mikko LAHTI; Maillefer, (Finland), janne.harjuhahto@maillefer.net, jaakko.harjuhahto@mailletech.net, mikko.lahti@maillefer.net
Jussi HANHIROVA; Aalto University (Finland), jussi.hanhirova@aalto.fi
Björn SONERUD, Verdilink Consulting AB, (Sweden), bjorn.sonerud@verdilink.se

## *ABSTRACT*

*We present a novel topography scanning system developed for XLPE cable core monitoring. Modern measurement technology is utilized together with embedded high-performance computing to a build complete and detailed 3D surface map of the insulated core. Cross sectional and lengthwise geometry errors are studied, and melt homogeneity is identified as one major factor for these errors. A surface defect detection system has been developed utilizing deep learning methods. Our results show that convolutional neural networks are well suited for real time analysis of surface measurement data enabling reliable detection of surface defects.*

## INTRODUCTION

Modern trends towards smart manufacturing principles require more meaningful data from production processes. Constant monitoring of production quality is a key part in moving towards smarter process controls. During the power cable insulation process, the cable core is traditionally inspected by mainly diameter gauges and rarely by any precise shape or surface monitoring system. Furthermore, surface quality is monitored visually, and the inspection is not automated. Employing automated quality monitoring systems facilitates the use of machine learning methods as a part of process control and enables, e.g., the possibility for predictive maintenance as soon as slight variations in the product are measured.

A power cable insulation process consists of many phases, of which each have significance for the polymeric cable core quality. During the extrusion process each of the material layers are melted, mixed and distributed evenly around the circumference of the conductor. The used polyethylene materials contain peroxides, which are thermally unstable molecules, in order to crosslink the polymer chains. The crosslinking creates a network of polymer chains which allows for a higher operating temperature. Any issues during the extrusion process may lead to premature crosslinking, i.e. scorch, inside the extruder or crosshead, causing quality problems.

During the next phase, curing and cooling processes, the cable core is subject to high temperatures and undergoes significant thermal expansion simultaneously with the crosslinking process, and shrinking during subsequent cooling.

The thermal expansion and cooling cause changes in the core geometry, which are apparent in the core shape after the process. This process can cause cross sectional non-circularity, i.e. ovality, or flat areas at some parts of the core. Furthermore, inhomogeneous melt quality is visible after curing as local waviness. This is not to be confused with lengthwise diameter variation caused by the rotations of the screw. Diameter variation is typically across the whole cross section, whereas inhomogeneous melt appears as roundish bumps and indentations.

Neither of these geometrical issues are measured in typical high or extra high voltage production lines. We present a novel solution for measuring these geometrical quality metrics, a topography scanner measuring the core surface with laser displacement sensors. These quality metrics may be used towards smarter process control and for minimizing shape and diameter variations.

Optimizing core geometry has many benefits from insulation material saving to easier and better-quality joint and termination interfaces. We have conducted experimental work on modelling the geometrical changes as functions of process parameters. A wide range of trials have been conducted at a pilot vertical line at various conditions and with different constructions.

It has been shown previously, that the homogeneity of the extruded insulation material has significance for the dielectric strength of the core [1]. It is well known that there are remaining uneven mechanical stresses in the XLPE insulation [2]. Various studies have found decreased dielectric strength in specimens with residual stresses or mechanical strain [3-6].

On top of geometrical errors, the cable core may also be subject to local surface defects, such as scorched material embedded into the insulation or surface scratches from unintended contact with manufacturing equipment. We employ a machine learning based defect detection system to alert the manufacturer of any surface defects. The system is trained with purposefully manufactured defects made at a pilot vertical line, as well as with training samples received from cable manufacturers. Deep convolutional neural networks are shown to work well for this application, and we present key metrics of performance.

## TOPOGRAPHY SCANNING SYSTEM

Our experimental setup is built upon a power cable surface scanning instrument that we have developed. This device was designed for measuring the surface geometry of power cable products in real-time in order to establish online quality metrics for the product and the production process. The device generates a detailed 3D mesh of a cable's surface geometry: a topographic map of the surface. This surface map is then used to compute analytic quality metrics for the product and for detecting the presence and type of surface defects, such as incisions, scratches, extrusion residues, scorched raw material, etc.

This topography scanning system is composed of several optical sensors, a control and computing unit and a defect detection system. Figure 1 presents a 3D render of the topography scanner installed as a part of a power cable production line.

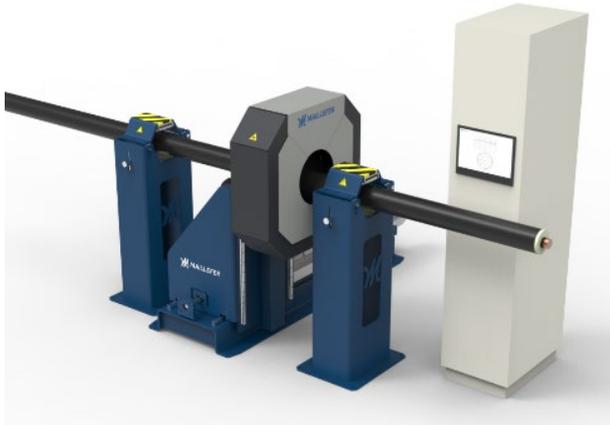

**Figure 1. Topography scanning system as part of a manufacturing line**

## Sensor subsystem

An array of laser displacement sensors is placed around the continuously moving cable product in an arrangement to cover the full 360° of the cylindrical cable. Each sensor measures the distance from itself to an arc of the cable's circumference. Merging these measurements into a single cross section yields a single profile and a sequence of profiles forms the surface mesh. For a typical high-voltage power cable application, the sensors operate at a sampling rate of 200 Hz to 800 Hz while the cable moves at a speed of 0.5 - 2.0 meters per minute.

## Measurement subsystem

The sensor system is attached to a server running a custom data acquisition and processing application. Measurements from individual sensors are assembled into profiles and the data is filtered and conditioned as necessary.

The number of raw data points per profile depends on the number of sensors, the size and position of the measured sample. During regularization, profile size is reduced to a fixed number of points e.g. 3600, equally spaced around the sample to facilitate further processing and analysis, such as computing quality metrics for the product.

The computing server offers an interface for publishing the full profile data and any derived metrics to subscribers over a local network. The three most significant subscribers are: a visualization client for real-time monitoring of metrics and the cable surface, a SCADA plant automation systems for logging and alerting, and the defect detection system used for identifying abnormalities on the cable products surface.

## Metrics

From the detailed cross section shown in figure 2, we calculate the minimum diameter $D_{min}$ and maximum diameter $D_{max}$. From these, we derive the absolute ovality [eq 1.] in millimetres and the relative roundness [eq 2].

$$D_{max} - D_{min} \quad [1]$$

$$\frac{D_{min}}{D_{max}} \quad [2]$$

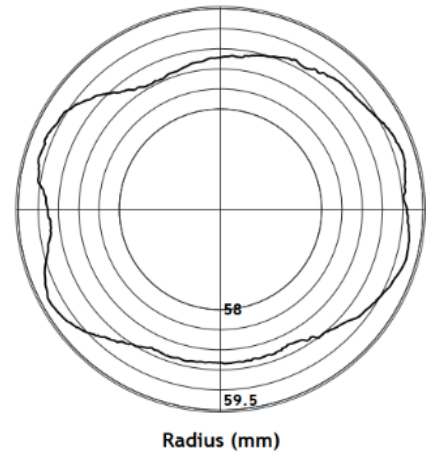

**Figure 2. Example of a cross sectional profile of an EHV cable sample showing 1.3 mm ovality.**

To visualize the 3D surface map of the cylindrical cable, we convert the surface map to polar coordinates and represent the surface as a plane in, as shown in Figures 3 and 4. Figure 3 represents a cable with low lengthwise surface variation and moderate ovality while figure 4 represents a cable with more lengthwise variation of the surface.

We introduce the concept of measuring local lengthwise variation of the surface, e.g., the waviness of the surface. We have verified that this waviness is due to the insulation layer and not the overlaying semi conductive layer.

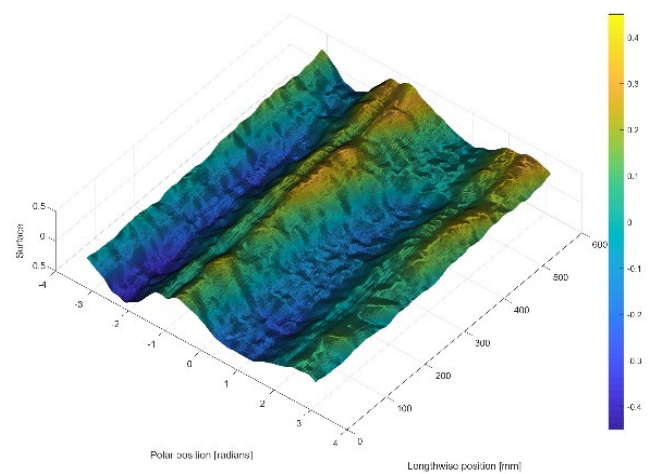

**Figure 3. Topography map of a cable sample with low lengthwise variation. X axis is polar coordinate (radians), y lengthwise position (mm) and Z surface deviation from ideal cylinder shape (mm).**

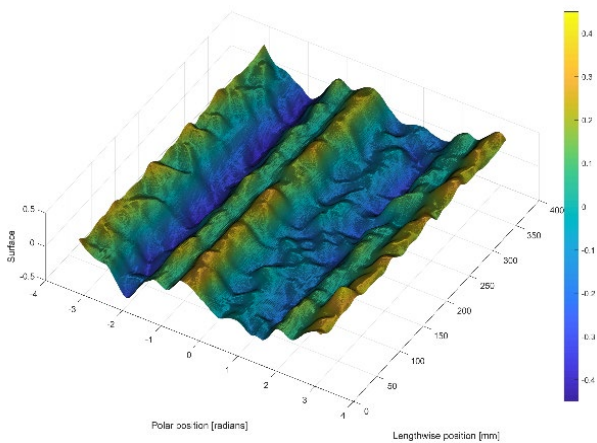

**Figure 4.** Topography map of a cable sample with high lengthwise variation. X axis is polar coordinate (radians), y lengthwise position (mm) and Z surface deviation from ideal cylinder shape (mm).

## A PILOT VERTICAL VULCANIZATION SETUP

The pilot vertical vulcanization setup is a unique platform for insulated power cable development capable of producing insulated cable cores at realistic processing conditions. Figure 5 shows the pilot vertical vulcanization setup and it's components. The extrusion group consists of 60 mm, 200 mm and 80 mm extruders for the inner semiconducting layer, the insulation and the outer semiconducting layer, respectively. Multiple triple crossheads used for these trials are typical for high voltage and extra high voltage lines up to 180 mm core outer diameter. The curing tube is 4.5 meters long pressure vessel which is heated with resistors and pressurized with nitrogen up to 10 bars.

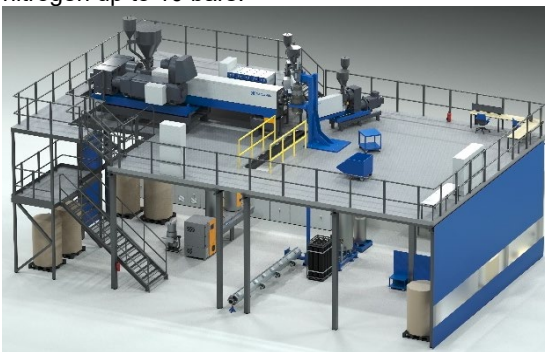

**Figure 5.** Pilot vertical vulcanization setup and components.

A model cable sample production proceeds along the following steps. First, the extrusion group is prepared, crosshead installed and heated up to process temperature. Second, the extrusion of all three plastic layers is started. Third, after stabilization of the extruders, the curing tube is lifted to upright position and heated up to starting temperature. Fourth, the sample production is started, which means the conductor will start moving at line speed and the curing tube is connected to the crosshead and pressurized. Fifth, the line is run at realistic production speeds for the whole length of the sample, after which the line stops, and the extruder group starts cooling to a stand-by temperature profile. Sixth, the curing process is now simulated by changing the curing tube temperature according to the curing recipe. Contrary to a production line, the model cable is not moving, but the temperature profile is adjusted to achieve the same thermal history as if the cable would be continuously moving through the curing tube. Finally, after the curing process is complete the sample is cooled with water and the curing tube is depressurized, the cable cut and removed from the curing tube.

This process combines all the main aspects of CV process, extrusion, curing and cooling. With this combination of processes, it is possible to study the interactions and the results of each aspect both on their own and in combination with others. The trial results have been verified by comparing them with cable samples from VCV lines with identical process conditions. All the comparisons have resulted in nearly identical results, proving that the pilot vertical line is a reliable tool for development.

## EXTRUSION AND CV PROCESS PARAMETERS

Process parameters of both extrusion and CV processes affect the final product. We have made over 55 trials in order to model these interactions mainly for VCV lines, but same underlying mechanisms are present in CCV lines. Samples in this study range from 60 mm to 150 mm in outer diameter. We have constructed a mathematical nonlinear model using measured geometry data. As it is not feasible to test all possible variations between the parameters, our process model is an advantageous tool for highlighting the effects of each individual parameter.

For the extrusion process, the extruder and screw combination is the most significant factor affecting the melt quality and extrusion conditions have a secondary effect. We have worked through multiple development cycles for the insulation screw designs in order to maximize output while minimizing inhomogeneity of the melt. Figure 6 shows how the screw geometry is more dominant factor than extrusion conditions, when melt homogeneity is studied via lengthwise waviness of the surface.

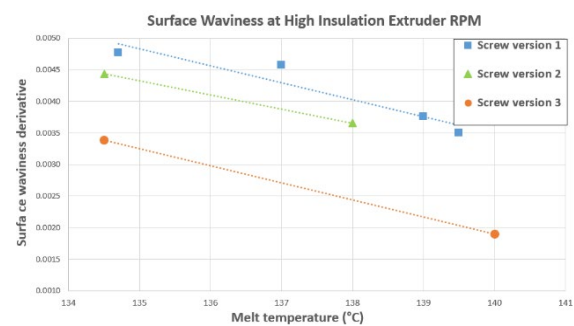

**Figure 6.** Surface lengthwise waviness measured from crosslinked cable samples. Three versions of insulation screw were evaluated at various process conditions at the same high insulation extruder RPM.

The surface waviness illustrated in figures 4 and 5 is more pronounced when running the extruder at maximum speed and with harsh CV conditions. We propose that insulation melt inhomogeneity leads to uneven behaviour during crosslinking and thermal expansion, which is why it is visible only after curing. A core with high lengthwise surface waviness will likely contain higher residual stresses than a smooth core. Inhomogeneity of the insulation melt as itself has been shown to have negative effect on dielectric strength [2] and combined with potentially higher residual stresses raise concern about decreased dielectric strength.

Development done regarding melt homogeneity aims towards increasing insulation extruder output. While operating at a typical RPM range of 15-20 RPM for a 200 mm extruder the issue is quite limited. Issues start to arise when increasing the screw RPM closer to a maximum of a given extruder size. Residence time of insulation material inside the extruder will be shorter and it is more challenging to reach a sufficient level of melting and mixing.

Cross sectional deviation from perfect roundness is strongly influenced by crosshead design, but it is not independent of other process parameters. Melt homogeneity and other process parameters have significant influence on the roundness of the core as shown in Figure 7 and on lengthwise variation as shown in Figure 8. Bulsari *et al.* [7] presents the relationships and discuss modelling work done on this subject.

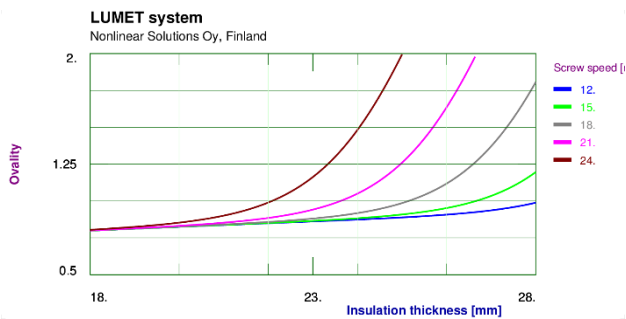

**Figure 7. Effect of insulation thickness on ovality for different values of insulation extruder rpm showing smallest ovality for lowest screw speed and highest ovality for fastest screw speed. [7]**

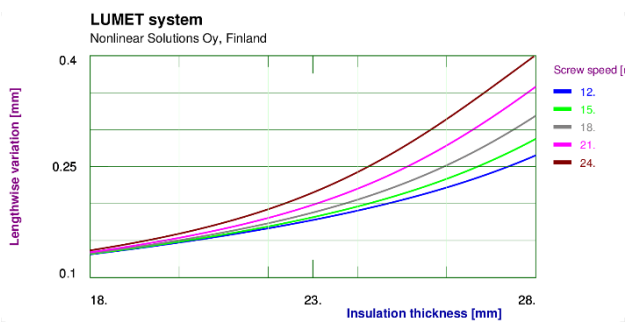

**Figure 8. Effect of insulation thickness on lengthwise variation for different values of insulation extruder rpm showing smallest lengthwise variation for lowest screw speed and highest lengthwise variation for fastest screw speed. [7]**

Having a model for ovality and lengthwise surface waviness will help with improving smart manufacturing features of CV lines. Cable core surface scanning combined with a software model can present the line operator with an expectation of quality combined with real-time measurement of the quality. We can identify issues with malfunctioning equipment, unstable process variables and see whether the line is in need of immediate maintenance.

## DEFECT DETECTION SYSTEM

The defect detection system utilizes convolutional neural networks (CNN) to detect surface defects in the measurement subsystem data. The defect detection system retrieves profile measurement data from the measurement subsystem, sends the retrieved data for classification to the inference subsystem, and finally returns the classification results to the topography scanning measurement system.

### Convolutional neural networks

CNNs consist of model architecture and model weights. Model architecture defines the structure of the computation, I.e., number and type of internal operations. Model weights are iteratively acquired numerical variables attached to the CNN internal operations. They are iteratively acquired from CNN training, and represent the "knowledge" of the CNN.

The CNN model internal weights are generated by iteratively training the model using human-annotated training data. A trained CNN model can be used for inference, i.e. to predict the class of unseen data.

CNNs have been extensively used in classification of image data [8]. The topography scanner 3D surface data can be projected onto 2D grayscale image data. This allows evaluation of well-known CNN models for analysis of topography scanner data.

In order to train a CNN model for defect detection we have created a training data library. We have measured multiple cable samples using topography scanning, and project the data onto grayscale image data. The measured samples contain real and artificially generated defects, which we have annotated. Figure 9 shows the distribution of annotated defects per size category. The total size of our training data library is over 8 billion measurement points. From the training library we can generate different training datasets, with varied augmentation operations.

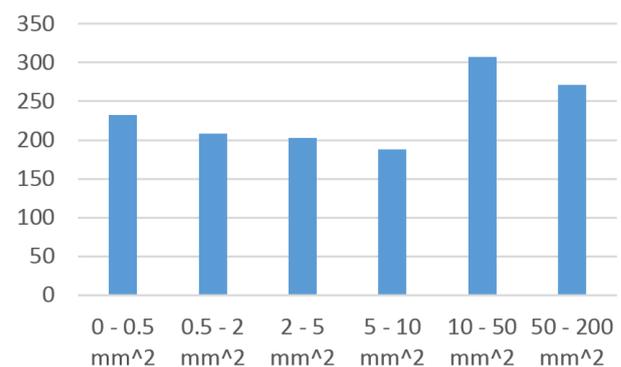

**Figure 9. Number of annotated defects per size category in our training data library.**

Topography scanning produces a 3D surface mesh of the defects. Examples of surface defects are shown in figures 10 and 11. As the dimensions of the 3D surface mesh are known, it is possible to calculate the exact size of interesting features, such as surface defects.

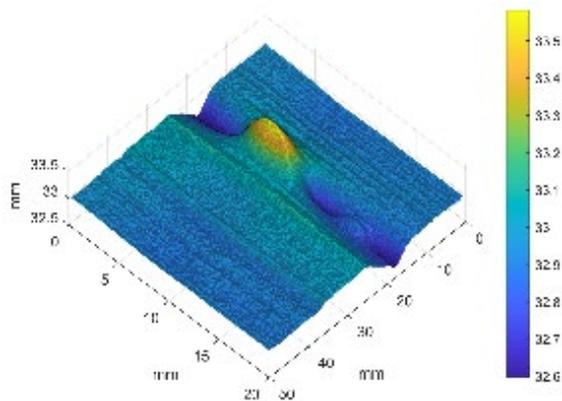

**Figure 10. Industrial sample with scorch of outer semicon layer. Defect width 9 mm, length 5 mm, heigth 0.4 mm.**

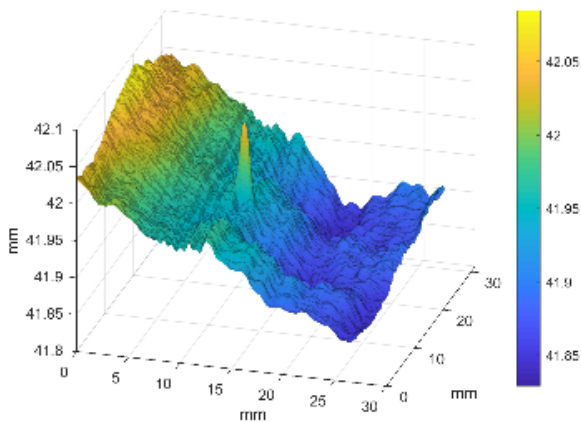

**Figure 11. Example of a small scorch particle. Defect width 0.6 mm, length 0.7 mm, heigth 0.06 mm**

In order to evaluate the viability of the CNN based approach to defect detection, and to demonstrate the functionality, we have chosen two well-known open source CNN models. The CNN models we use are the MobileNet and the ResNet-50. [9] The models are pre-trained with the ImageNet dataset [10], and we use transfer learning [11] to fine-tune the CNNs for classifying cable surface data.

A CNN inference result gives the probability for each input data sample belonging to a certain class. The classification taxonomy is defined when preparing and training a CNN model. For example, a minimal classification taxonomy consists of two labels: "clean" and "not-clean", the latter indicating that the surface patch contains a defect, scratch or some other type of surface aberration.

Classifying surface defects using a more complex taxonomy is beneficial for determining the type and possible root cause of each defect. As an example the surface defects shown in figure 12 can be classified into two different classes. Subfigures 12.a and 12.b are examples of scorch during production whereas 12.c and 12.d are results of mechanical contact.

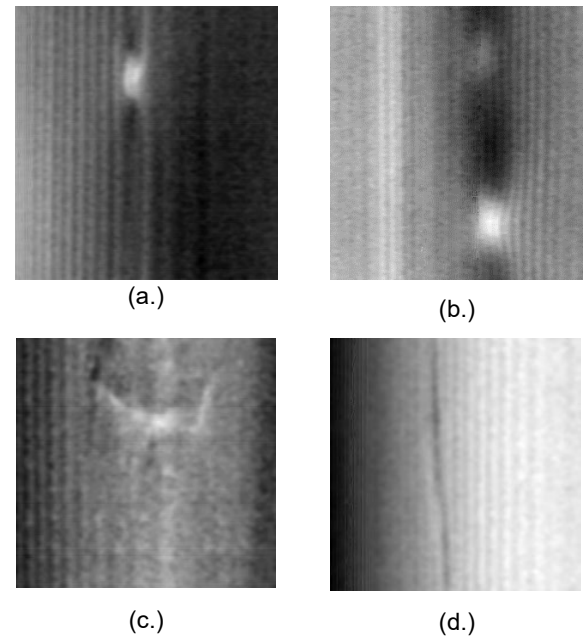

(a.)    (b.)

(c.)    (d.)

**Figure 12. Greyscale rendering of surface map with local surface defects: (a) small scorch particle of outer semicon, (b) large scorch particle of insulation at outer semicon interface, (c) surface damage from unintented contact, (d) lenghtwise scratch on surface.**

We evaluate the inference subsystem classification using two metrics. First, does the system detect all of the defects. Second, how much it generates false positives.

### Two-phase defect detection

In many industrial deployments a low false positive rate is required. At the same time a high defect detection rate is desired. We have created a two-phase defect detection system that has internal parameters to control the detection, so that it can be tuned to meet the requirements of both false positive rate minimization and maximization of the defect detection rate. The internal parameters also allow fine tuning the detection system to process specific needs or customer requirements. Configurable parameters include, e.g., threshold values for detection probabilities that trigger defect detection alarms, possibilities for weighting the model detection capability towards certain defect classes, or optimization of the detection window size for certain special use-cases. Flowchart of the two-phase defect detection logic is presented in figure 13.

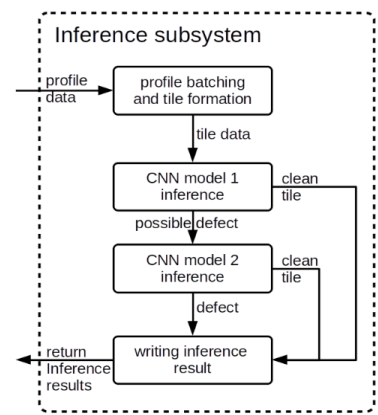

**Figure 13. Flowchart of defect detection system.**

The defect detection system retrieves profile measurement data from the measurement subsystem and maintains a batch buffer for the profile data. Based on overlap configuration, it transforms the profile data batches into overlapping tiles that are sent for inference. Overlapping tiles ensures that each point on the sample's surface is analysed several times for maximizing detection rate. Results of the CNN inference are transmitted back to the measurement subsystem.

To evaluate the detection performance of our two-phase detection system, we define a custom defect hit ratio metric. The defect hit ratio measures how many of the defect areas on a cable sample are detected. We also evaluate the performance by measuring the false positive rate of the detection system. False positive rate measures how many false alarms the system generates.

We measure the abovementioned metrics using a test dataset consisting of data from cable samples not used in the CNN training process.

Results of the two-phase detection performance evaluation are presented in Figure 14. On the x-axis is the false positive rate and on the y-axis the defect hit ratio. Results are presented for both the individual CNNs and their combined two-phase performance. Measurement points with numerical labels present different overlap percentage value configurations for the system.

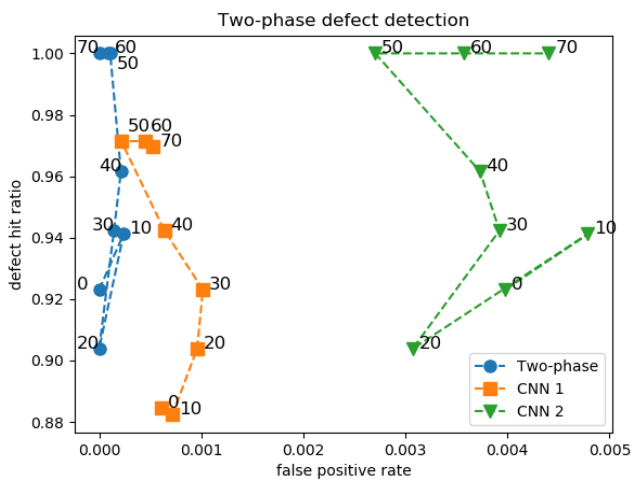

**Figure 14. Defect hit ratio and false positive rate of two separate CNNs and a two-phase system.**

Results presented in Figure 14 show that with the test dataset our two-phase detection system is able to find the defects with an accuracy of 100%, while keeping the false positive rate at 0%. Figure 14 also shows that the combined two-phase detection performs better than either of its components individually.

As real defects are difficult to gather, and because all production lines differ slightly on their product quality details, fine-tuning of the defect detection system can be continued on-site. On-site tuning also includes configuring the detection system parameters according the cable manufacturers requirements

## CONCLUSIONS

Topography scanning provides new cable geometry measurement information, which has not been previously accessible in real time.

Experimental work shows strong correlation between geometrical irregularities of the cable core, extrusion equipment and process parameters. Understanding the relationships affecting core geometrical quality improves the cable insulation process and guides machinery optimization.

Further work is recommended to verify if inhomogeneity of insulation melt after extrusion increases residual stresses and decreases dielectric strength.

A deep learning surface defect detection system is trained and evaluated using real data. CNNs are able to classify surface data and yield very high defect hit rate while maintaining near zero false positive rate.

**GLOSSARY**

**CNN:** Convolutional neural network
**CCV**: Catenary continuous vulcanization
**VCV**: Vertical continuous vulcanization
**SCADA:** Supervisory control and data acquisition
**XLPE:** crosslinked polyethylene